\documentclass[11pt]{article}

\usepackage[preprint]{acl}

\usepackage{times}
\usepackage{latexsym}
\usepackage{enumitem}

\usepackage[T1]{fontenc}
\usepackage{amsmath}
\usepackage{booktabs}
\usepackage{multirow}
\usepackage{hyperref}

\usepackage[utf8]{inputenc}

\usepackage{microtype}

\usepackage{inconsolata}

\usepackage{graphicx}
\usepackage{subcaption}
\usepackage{float}

\usepackage{xspace}

\newcommand{\agenticname}{{Deontic Agentic Reasoning}\xspace}
\newcommand{\abbrev}{{DAR}\xspace}

\title{\abbrev: Deontic Reasoning with Agentic Harnesses}

\newcommand{\aspace}{\hspace{1em}}
\newcommand{\jhu}{$^{\heartsuit}$}
\newcommand{\telecom}{$^{\spadesuit}$}

\author{%
    \textbf{Guangyao Dou}\jhu\aspace
    \textbf{William Jurayj}\jhu\aspace
    \textbf{Nils Holzenberger}\telecom\aspace
    \textbf{Benjamin Van Durme}\jhu\aspace\\[0.4ex]
    \jhu Johns Hopkins University\aspace\telecom T\'el\'ecom Paris, Institut Polytechnique de Paris\\[0.4ex]
}

\begin{document}
\maketitle

\begin{abstract}
Deontic reasoning is the task of answering questions by applying explicit rules and policies to case-specific facts, for example computing tax liability under a statute or determining the outcome of an immigration appeal. A key technical challenge for LLM-based deontic reasoning is that the relevant ruleset can be long and cross-referenced, so models may still fail to locate the rules needed for a particular reasoning step. We introduce \agenticname (\abbrev), an agentic reasoning setup in which the model interacts with the statutes on demand. We evaluate \abbrev under multiple harnesses on hard subsets of DeonticBench. Across these settings, we find that agentic harnesses can push the frontier on deontic reasoning tasks, but improvements are not uniform: weaker models often degrade on numerical tasks while consuming far more tokens. Code is available \href{https://guangyaodou.github.io/harbor-deonticbench/}{here}.

\end{abstract}

\begin{figure*}
    \centering
    \vspace{-1.2ex}
    \includegraphics[width=0.88\textwidth]{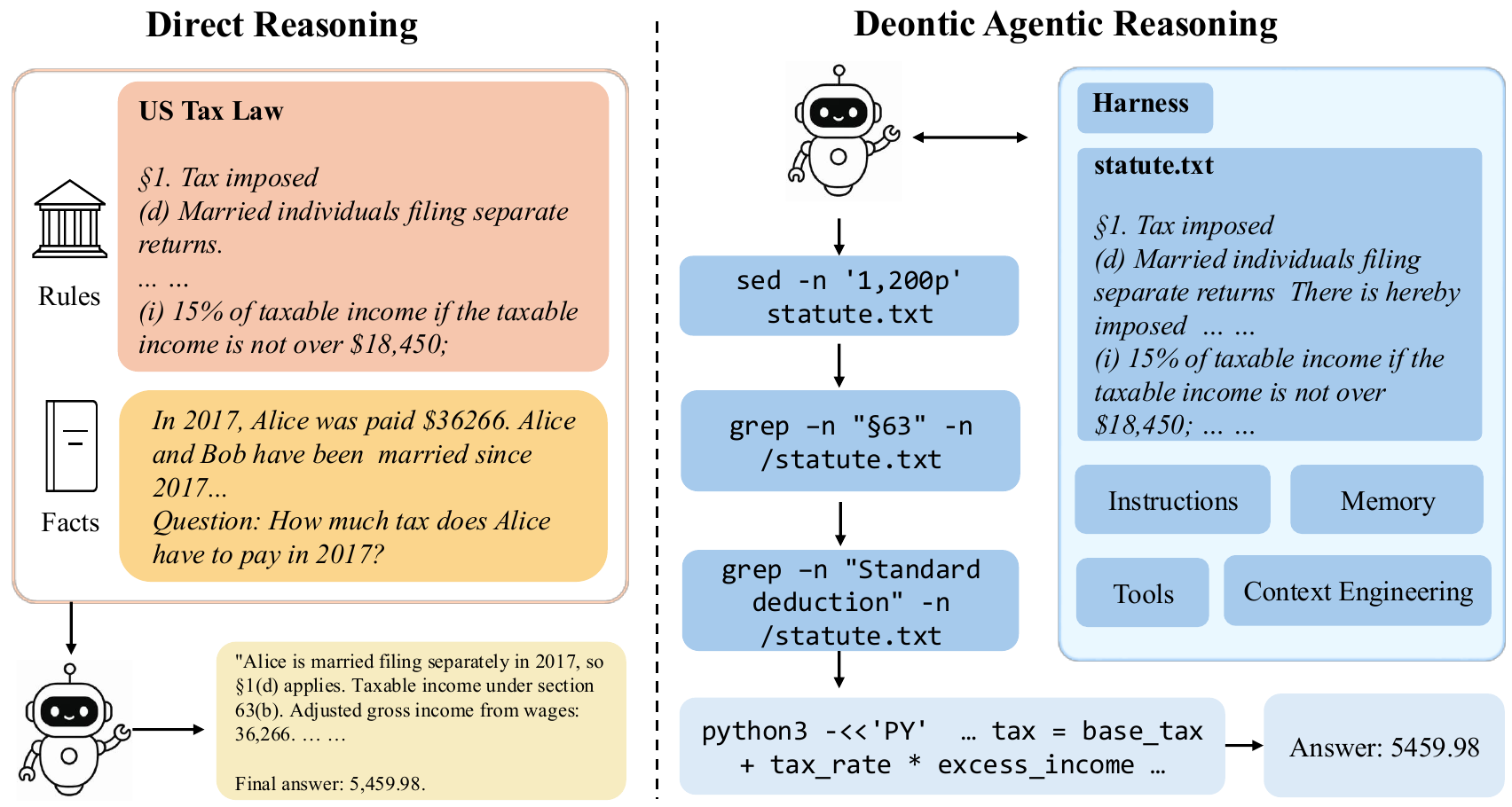} 
     \caption{Direct reasoning vs. \agenticname (\abbrev). In direct reasoning (left), the full statute and case facts are placed in the prompt, and the model produces an answer in a single pass. In \abbrev (right), the statute is placed as a file in the harness, and the model examines it on the fly using general-purpose tools.}
\label{fig:isr-overview}
\end{figure*}

\section{Introduction}

Deontic reasoning, the task of answering questions by applying explicit rules and policies to case-specific facts, is a core capability for language models deployed in high-stakes domains such as tax computation \citep{holzenberger-van-durme-2023-connecting} and policy compliance \citep{zhou2025rulearena}. The technical difficulty is the rulesets themselves: statutes are long and heavily cross-referenced, with most provisions irrelevant to any given case and obligations qualified by definitions and exceptions located elsewhere in the text.

The standard setup for evaluating deontic reasoning places the entire set of rules, case facts, and question in a single prompt, asking the model to find and apply the relevant rules in one pass \citep{dou2026deonticbench, jurayj2026language, zhou2025rulearena}. Yet recent work on agentic search shows that on factual retrieval tasks, models handle long corpora better when they search them with general-purpose tools (grep, file reads, shell commands) than when they receive them as static context \citep{li2026beyond, sen2026grep}. Whether the same is true for deontic reasoning, where the task is not retrieval but reasoning grounded in rules, is an open question.

We study this question by introducing \agenticname (\abbrev), a setup in which the statute is placed as a file in a harness environment and the model examines it on demand. We evaluate \abbrev on DeonticBench \citep{dou2026deonticbench}, where each task consists of a statute, a case fact, and a question. The benchmark covers U.S. federal tax (SARA), U.S. immigration administration (USCIS), and airline baggage policies (Airline).

Our experiments across frontier and open-source models show that \textbf{agentic harnesses improve frontier models on the deontic reasoning tasks but degrade weaker models on the same tasks.} For frontier models, the harness enables self-directed retrieval and lets models recover from intermediate errors. For weaker models, the same scaffolding becomes a confidence amplifier, spending more tokens on the same wrong answer instead of intelligently stopping early \cite{wang2026conformal}. On SARA-Numeric, frontier models gain 15–30\% under the Terminus-KIRA \citep{terminuskira2026}, while open-source models in the same harness degrade by 11–23\%. This suggests that a harness gives the model interactive access with tool use, but not the underlying judgment to use it well. Our main contributions are threefold:
\begin{itemize}[leftmargin=*, itemsep=2pt, topsep=2pt, parsep=0pt]
\item We present \agenticname, a setup in which deontic reasoning agents access statutes on demand through a harness rather than receiving them in context.
\item We perform a systematic comparison of \abbrev against direct prompting on DeonticBench, spanning frontier and open-source models.
\item We show that \abbrev's effect depends on model capability. Under Terminus-KIRA, frontier models gain 18–30 points on SARA-Numeric. The same harness \emph{degrades} weaker open-source models: Qwen3.5-35B drops from 34\% to 11\% on SARA-Numeric, and every open-source model collapses to near-zero on Airline while consuming up to  $4\times$ more tokens per trial.
\end{itemize}

\begin{figure*}[t]
    \centering
    \includegraphics[width=0.5\textwidth,trim=0 6 0 6,clip]{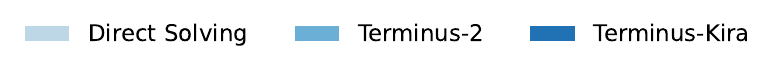}

    \vspace{-0.3em}

    \includegraphics[width=0.495\textwidth,trim=0 10 0 10,clip]{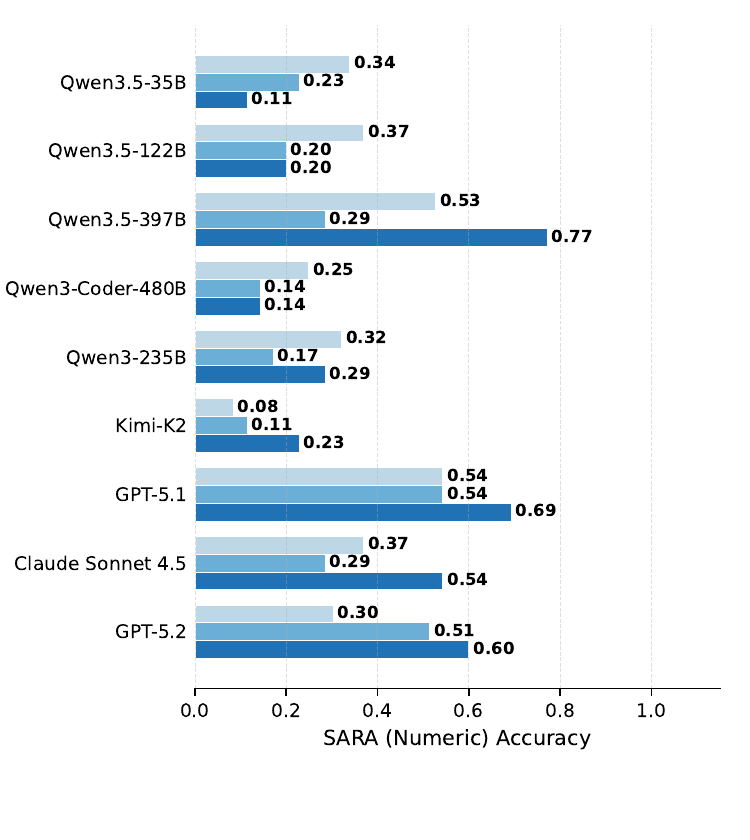}\hspace{-0.03\textwidth}%
    \includegraphics[width=0.495\textwidth,trim=0 10 0 10,clip]{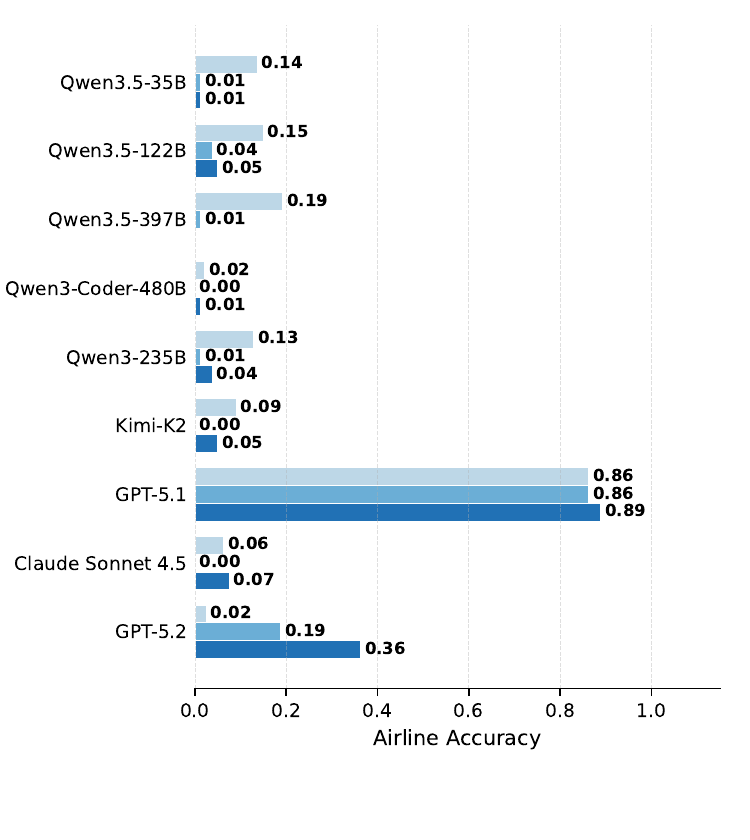}

    \vspace{-1.7em}

    \includegraphics[width=0.495\textwidth,trim=0 10 0 10,clip]{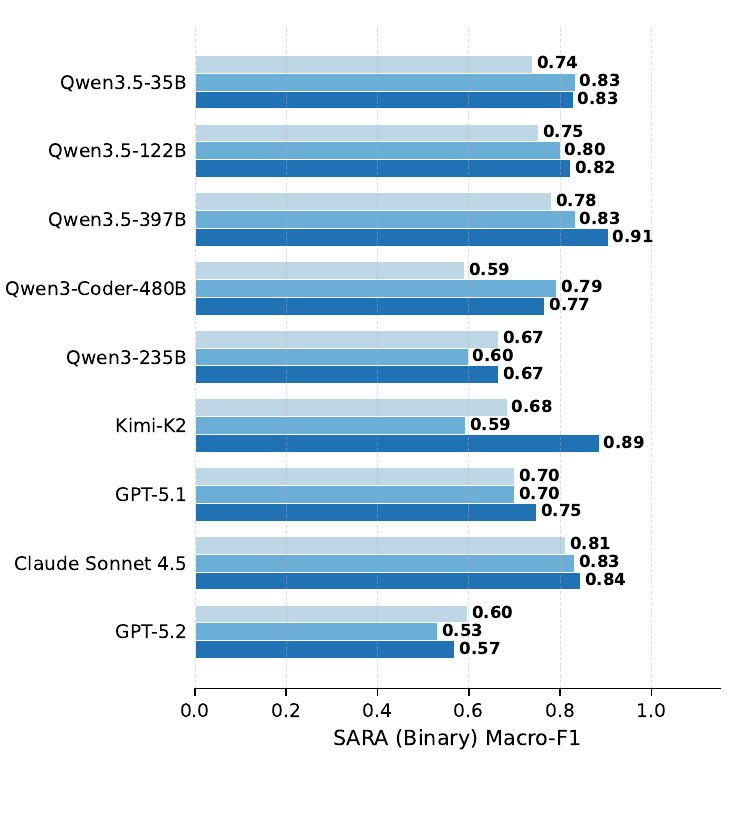}\hspace{-0.03\textwidth}%
    \includegraphics[width=0.495\textwidth,trim=0 10 0 10,clip]{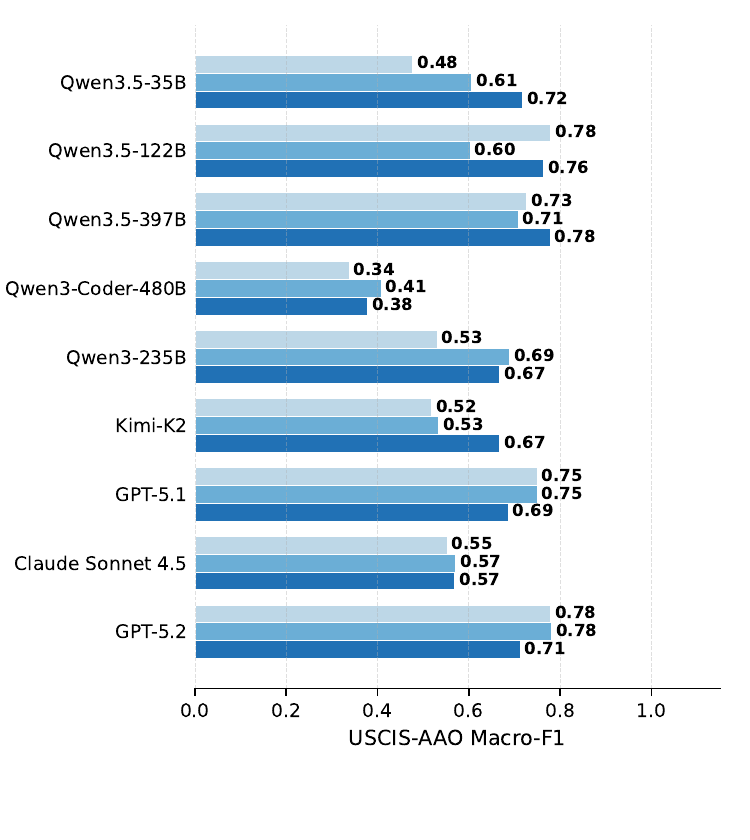}

    \vspace{-1.8em}
    \caption{Harness comparison across DeonticBench.}
    \label{fig:combined-models-nosubcap}
\end{figure*}

\section{Deontic Agentic Reasoning}
\label{sec:method}

We compare two paradigms for deontic reasoning over statutes, illustrated in Figure~\ref{fig:isr-overview}.

\paragraph{Direct reasoning.} In this setup, the model receives the full statute, the case facts, and the question in a single prompt, and produces an answer in one pass. This is the configuration used in most prior deontic reasoning evaluations \citep{dou2026deonticbench, jurayj2026language, zhou2025rulearena}. The statute is fully present in context, and the model must identify the applicable provisions and reason over the entailed obligation in a single inference.

\paragraph{\agenticname (\abbrev).} In \abbrev, the statute is not part of the prompt and placed as a file (\texttt{statute.txt}) in a harness environment. The model receives the case facts and the question, along with instructions describing the harness and its tools. To answer the question, the model issues tool calls to read targeted portions of the statute on demand. In a simple terminal-based harness, these include shell commands such as \texttt{sed}, \texttt{grep}, and \texttt{cat}. The model may issue arbitrarily many tool calls and may also execute Python for numeric computation. Each tool call produces an observation that is appended to the context for subsequent turns, so the agent accumulates observations as it explores. We use the term \abbrev to emphasize that the model interacts with the statute as a queryable resource rather than receiving it as static context.

\section{Experimental Setup and Results}

\subsection{Datasets}

We evaluate on DeonticBench \citep{dou2026deonticbench}, a suite of deontic reasoning tasks drawn from legal and airline baggage-policy domains \citep{zhou2025rulearena}. Each task consists of a statute, a case fact, and a question. In this work, we focus on the hard subset of DeonticBench:

\begin{itemize}[leftmargin=*, itemsep=2pt, topsep=2pt, parsep=0pt]
    \item \textbf{SARA (Numeric):} numerical tax-liability computation, evaluated by accuracy.
    \item \textbf{SARA (Binary):} binary statutory-entailment classification, evaluated by macro-F1.
    \item \textbf{Airline:} application of airline-passenger baggage fee policies, evaluated by accuracy.
    \item \textbf{USCIS-AAO:} immigration-appeal outcome prediction, evaluated by macro-F1.
\end{itemize}

\subsection{Models}
We evaluate nine models spanning open-source and proprietary models. For open-source models, we test three sizes of the Qwen3.5 family \citep{qwen3.5}: Qwen3.5-35B-A3B, Qwen3.5-122B-A10B, and Qwen3.5-397B-A17B. We also evaluate Qwen3-Coder-480B, Qwen3-235B-A22B \citep{qwen3}, and Kimi K2 0905 from moonshot \citep{kimiteam2025kimik2openagentic}. For proprietary models, we evaluate OpenAI GPT-5.1 and GPT-5.2 \citep{singh2025openai} with reasoning effort set to none, and Claude Sonnet 4.5 \citep{anthropic2025sonnet45}. Open-source models are served via vLLM \citep{kwon2023efficient} or accessed through the OpenRouter API.

\subsection{Harness}
\label{setup-harness}

Harness execution is orchestrated by the Harbor framework \citep{Harbor_Framework}. Our main experiments compare direct solving against two agentic harnesses: Terminus-2 \citep{merrill2026terminalbenchbenchmarkingagentshard} and Terminus-KIRA \citep{terminuskira2026}. Terminus-2 is a terminal-based agent harness in which a model operates autonomously inside a sandbox environment through an interactive \texttt{tmux} session. Terminus-KIRA is built on Terminus-2 and targets failure modes observed when models run under Terminus-2, including premature submission and poor self-evaluation. We describe detailed harness-level differences in Appendix~\ref{app:harness-details}.
In our setting, these harnesses let models interactively inspect the provided statutes rather than reasoning only under direct solving. 


\subsection{Results}
Figure~\ref{fig:combined-models-nosubcap} reports Direct Solving, Terminus-2, and Terminus-KIRA across nine models on the four DeonticBench tasks. Each task is allotted a 10-minute budget; trials that exceed this budget, fail to parse, or raise harness runtime errors are counted as incorrect in Figure~\ref{fig:combined-models-nosubcap}. We provide a detailed failure-mode breakdown in Appendix~\ref{appendix:error-analysis}.


\paragraph{Frontier models gain from \abbrev.}
The three proprietary models improve on the two numerical tasks once given a harness. Under Terminus-KIRA, GPT-5.2 climbs from 30\% to 60\% on SARA-Numeric; Claude Sonnet~4.5 rises from 36\% to 54\% on SARA-Numeric; and GPT-5.1 still picks up an additional 15 percentage points on SARA-Numeric and remains saturated near 0.86 on Airline. The pattern holds on the classification tasks: every frontier model is at or above its Direct baseline under at least one harness on SARA-Binary and USCIS-AAO. For frontier models, the harness turns latent statute-reading ability into delivered accuracy, exactly as the Mismanaged Geniuses Hypothesis~\citep{zhang2026mgh} would predict.

\begin{figure}[t]
    \centering
    \includegraphics[width=\columnwidth]{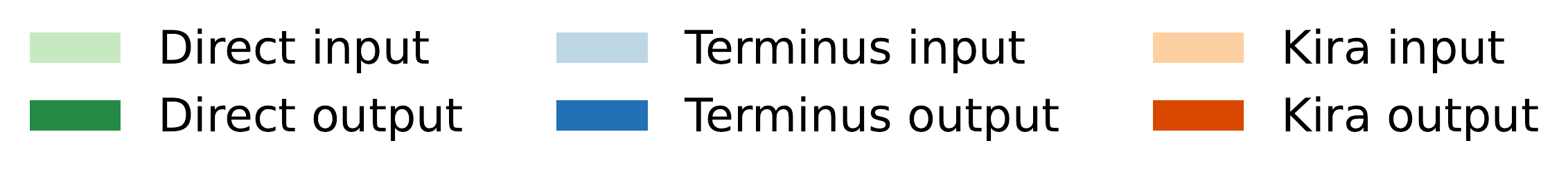}

    \vspace{-0.2em}

    \includegraphics[width=\columnwidth]{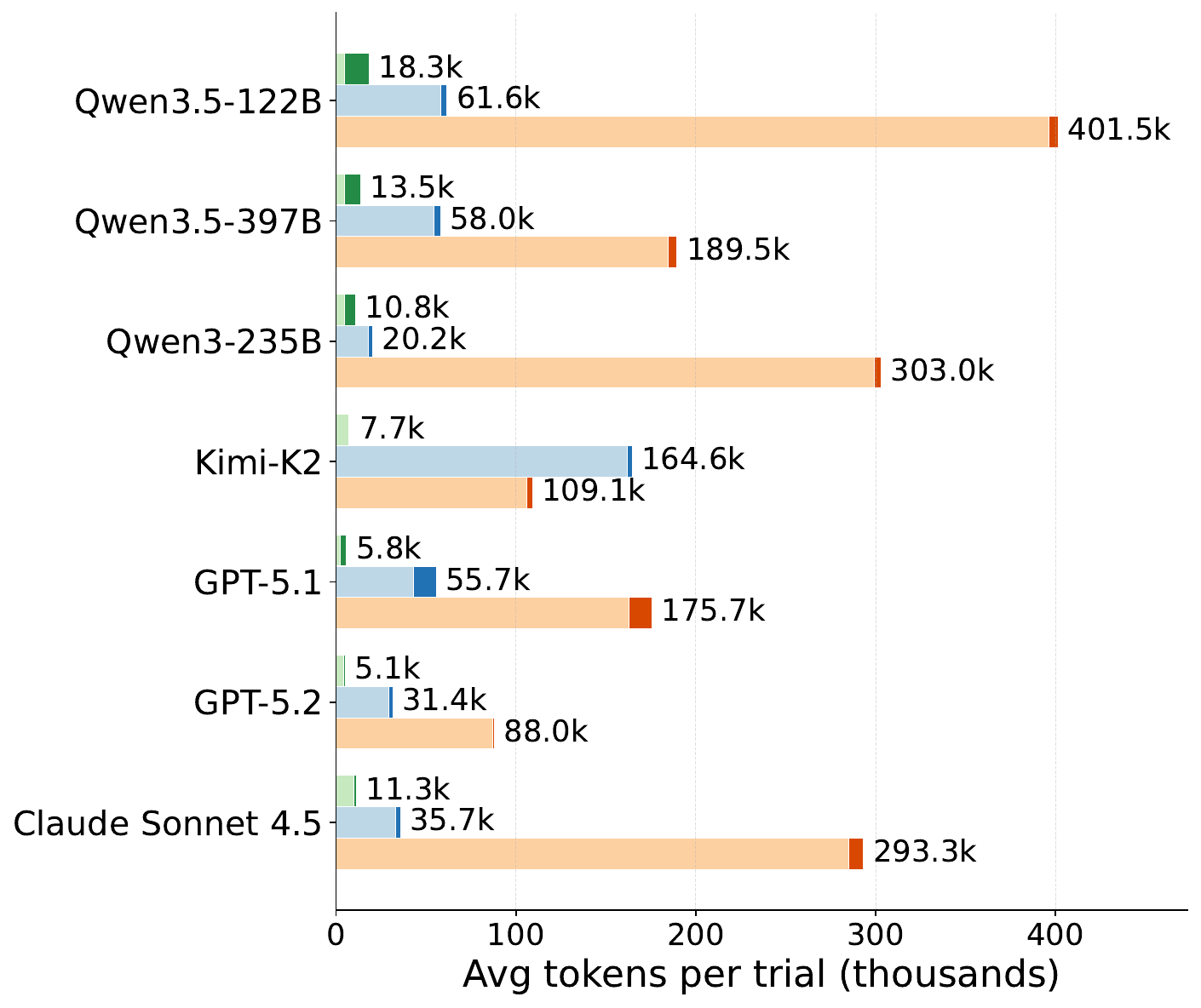}
    \caption{Average tokens consumed per trial under Direct Solving, Terminus,
    and Terminus-KIRA.}
    \label{fig:token-usage}
\end{figure}

\paragraph{Open-source models fail under the same harness.}
The same scaffold that helps the frontier hurts the open-source models, most severely on numerical reasoning. On SARA-Numeric, Qwen3.5-35B drops from 34\% to 11\% under Terminus-KIRA, Qwen3.5-122B from 37\% to 20\%, etc. The Airline panel is the cleanest illustration: every open-source model collapses to near-zero accuracy once placed in Terminus-2 or KIRA, even though their Direct baselines are non-trivial. Rather than enabling self-directed retrieval, the additional turns appear to inflate already-shaky reasoning into longer, more confident wrong answers. Figure~\ref{fig:token-usage} makes this concrete: under Terminus-2, Qwen3.5-122B averages 401k tokens per trial and Qwen3-235B 303k, roughly $4\times$ what the frontier models consume. The classification tasks show smaller harness-induced degradations but no consistent open-source gain that mirrors what the frontier obtains. Compared to direct solving, agentic harnesses consume more tokens because the output of each action is appended to the input of the next iteration, raising new challenges for balancing inference costs against answer utility \citep{jurayj2025your}.

\paragraph{Additional Experiments.}
In addition to Terminus-2 and Terminus-KIRA, we run experiments with Claude Code and Codex CLI. We report these additional results in Appendix~\ref{appendix:additional-cc-codex}. We also evaluate Recursive Language Models on DeonticBench \citep{zhang2025recursive}, with results reported in Appendix~\ref{appendix:additional-rlm}.

\section{Related Work}
\paragraph{Harness-based Agentic Search.} Prior work trains LLMs to interleave reasoning with search over a fixed retriever interface \citep{li2025flow, jin2025search, li2026openresearcher}. \citet{li2026beyond} invert this design with direct corpus interaction (DCI), letting the agent search the raw corpus using general-purpose terminal tools and showing large gains over conventional retrievers on agentic search and IR benchmarks. \citet{sen2026grep} report a complementary finding: lexical search over the corpus, paired with a capable harness, often outperforms semantic retrieval on long-memory QA. 

\paragraph{Deontic Reasoning Datasets.} Prior benchmarks focus on multi-step entailment and first-order-logic reasoning \citep{han2024folio, chen2025justlogic}. CL-bench \citep{dou2026clbenchbenchmarkcontextlearning} introduces context learning, testing whether a model can operate inside a rule system by following its rules. DeonticBench \citep{dou2026deonticbench} instead tests whether a model can reason from the outside about a specific case, grounded in the provided statute.

\section{Conclusion}
We introduce \agenticname and show that agentic harnesses push the frontier on the hardest deontic reasoning tasks, but improvements are not uniform: frontier models gain, while open-source models degrade and consume up to 4× more tokens. For sufficiently capable models, \abbrev unlocks the performance that static long-context prompts leave on the table.


\section*{Limitations}
\paragraph{Scalability of \abbrev.}
The current version of \abbrev places the entire statute as a single file in the harness and relies on the agent to navigate it through general-purpose tools like \texttt{grep} and \texttt{sed}. For the statutes in DeonticBench, this is tractable, but for substantially longer rulesets (e.g. the full U.S. Internal Revenue Code or large multi-jurisdiction regulatory corpora), even frontier models would need to read through large portions of the file to locate relevant provisions, consuming many tokens per case. A more scalable design would pair \abbrev with an efficient retrieval system, for example hierarchical statute lookup or learned section-level retrieval, that extracts relevant rulesets before the agent begins reasoning.
\paragraph{Benchmark and domain coverage.}
All of our results come from DeonticBench, which covers U.S. federal tax, U.S. immigration administration, and airline baggage policies. Real-world deontic reasoning spans many additional domains, including other section of laws and rule-following problems, each with structural properties. Replication on larger rule-grounded deontic reasoning benchmarks would strengthen the generality of our findings.
\paragraph{Harness coverage.}
We evaluate four harnesses: Terminus-2, Terminus-KIRA, Claude Code, and Codex CLI. The agentic harness space is moving quickly, and our results do not speak to harnesses designed specifically for statute reasoning, for instance with provision-aware navigation primitives or built-in cross-reference tools. Such a harness might change the capability-amplification picture for weaker models. One concrete path is automated harness search: Meta-Harness \citep{lee2026meta} discovers task-specific harnesses through outer-loop search over harness code and has surpassed hand-engineered baselines on agentic coding benchmarks. Applying it to \abbrev could surface statute-reading primitives tailored to deontic reasoning rather than relying on harnesses designed for general agentic tasks.

\paragraph{Reasoning-effort settings.}
We run GPT-5.1 and GPT-5.2 with reasoning effort set to none. Higher reasoning-effort settings may substantially change frontier performance and could narrow or widen the gap between frontier and open-source models that we observe.

\section*{Ethics Statement}

This work studies how language models reason over statutes and policies, using the publicly available DeonticBench benchmark, which includes questions from U.S. federal tax law, U.S. immigration administration, and airline baggage policies.

We highlight two ethical considerations relevant to our findings. First, deontic reasoning tasks such as tax computation and immigration appeal prediction involve high-stakes domains where model errors can carry real costs. Our results show that even with agentic harnesses, frontier models achieve only partial accuracy on these tasks, and weaker models can degrade further under the same scaffolding. We therefore caution against deploying current LLMs, with or without agentic harnesses, as autonomous decision-makers in legal, tax, or other high-stakes deontic contexts. The systems we evaluate are research artifacts and are not suitable substitutes for qualified human professionals.

Second, our finding that agentic harnesses act as capability amplifiers rather than universal fixes has implications for responsible deployment. Practitioners may assume that providing tool access to a model uniformly improves performance; our results suggest the opposite can hold for weaker models, where harnesses can produce more confident but less accurate outputs while consuming substantially more compute. This has both reliability and computational cost implications that should inform deployment decisions.

\section*{Acknowledgements}
This work was funded in part by the Defense Advanced Research Projects Agency (DARPA) CODORD program, and by Schmidt Sciences. Any opinions, findings, and conclusions or recommendations expressed in this material are those of the author(s) and do not necessarily reflect the views of DARPA or Schmidt Sciences.


\bibliography{custom}

\newpage
\appendix

\section{Harness details}
\label{app:harness-details}
Terminus-KIRA \citep{terminuskira2026} is built on Terminus-2 \citep{merrill2026terminalbenchbenchmarkingagentshard} and is motivated by failure-mode analysis of frontier models on Terminal-Bench. The blog post accompanying Terminus-KIRA identifies several patterns where the minimal Terminus-2 design lets capable models make avoidable mistakes:
\begin{itemize}[leftmargin=*, itemsep=2pt, topsep=2pt, parsep=0pt]
\item \textbf{Partial-work submission.} Models trained to assist humans tend to submit partial work rather than completing a task end-to-end. Terminus-2 does not actively counter this tendency.
\item \textbf{False completion.} When a model under Terminus-2 signals that it is done, the harness asks a single confirmation question, which models tend to answer affirmatively even when the task is incomplete or wrong.
\item \textbf{Brittle plan adjustment.} Models plan well from the initial information but struggle to revise their plans after observing new information mid-task.
\end{itemize}
Terminus-KIRA introduces harness-level changes intended to mitigate these patterns, particularly around completion verification and self-evaluation. In our deontic reasoning setting, these matter because many DeonticBench tasks have intermediate steps (locating a provision, applying a definition) that an over-eager model would skip past under a more permissive harness.

\section{Additional Results}
\label{appendix:additional-results}

\begin{table*}[t]
\centering
\renewcommand{\arraystretch}{1.25}
\setlength{\tabcolsep}{6pt}
\begin{tabular}{ll cc cc}
\toprule
& & \multicolumn{2}{c}{\textbf{Accuracy}} & \multicolumn{2}{c}{\textbf{Macro F1}} \\
\cmidrule(lr){3-4} \cmidrule(lr){5-6}
Model & Harness & SARA-Num & Airline & SARA-Bin & USCIS-AAO \\
\midrule
\multirow{5}{*}{Qwen3-Coder-480B}
  & direct        & 0.249          & 0.021          & 0.591          & 0.338 \\
  & codex         & 0.086          & 0.000          & 0.598          & 0.427 \\
  & terminus-2    & 0.143          & 0.000          & 0.793          & 0.408 \\
  & terminus-kira & 0.143          & 0.013          & 0.766          & 0.378 \\
  & claude-code   & \textbf{0.343} & \textbf{0.050} & \textbf{0.800} & \textbf{0.505} \\
\midrule
\multirow{5}{*}{Qwen3.5-122B}
  & direct        & \textbf{0.370} & \textbf{0.150} & 0.753          & \textbf{0.780} \\
  & codex         & 0.229          & 0.013          & 0.799          & 0.775 \\
  & terminus-2    & 0.200          & 0.038          & 0.800          & 0.603 \\
  & terminus-kira & 0.200          & 0.050          & \textbf{0.823} & 0.764 \\
  & claude-code   & 0.286          & 0.113          & 0.793          & 0.730 \\
\midrule
\multirow{4}{*}{Qwen3.5-35B}
  & direct        & 0.340          & \textbf{0.137} & 0.740          & 0.477 \\
  & terminus-2    & 0.229          & 0.013          & \textbf{0.833} & 0.607 \\
  & terminus-kira & 0.114          & 0.013          & 0.829          & \textbf{0.718} \\
  & claude-code   & \textbf{0.371} & 0.088          & \textbf{0.833} & 0.603 \\
\midrule
\multirow{4}{*}{Qwen3.5-397B}
  & direct        & 0.528          & \textbf{0.192} & 0.782          & 0.727 \\
  & terminus-2    & 0.286          & 0.013          & 0.833          & 0.708 \\
  & terminus-kira & \textbf{0.771} & 0.000          & \textbf{0.906} & \textbf{0.778} \\
  & claude-code   & 0.514          & 0.100          & 0.889          & 0.643 \\
\midrule
\multirow{4}{*}{Qwen3-235B}
  & direct        & \textbf{0.321} & \textbf{0.128} & 0.665          & 0.531 \\
  & codex         & 0.114          & 0.000          & \textbf{0.721} & 0.509 \\
  & terminus-2    & 0.171          & 0.013          & 0.598          & \textbf{0.689} \\
  & terminus-kira & 0.286          & 0.038          & 0.665          & 0.668 \\
\midrule
\multirow{4}{*}{Kimi-K2}
  & direct        & 0.084          & \textbf{0.090} & 0.684          & 0.518 \\
  & codex         & 0.200          & 0.000          & 0.733          & 0.553 \\
  & terminus-2    & 0.114          & 0.000          & 0.593          & 0.533 \\
  & terminus-kira & \textbf{0.229} & 0.050          & \textbf{0.885} & \textbf{0.668} \\
\midrule
\multirow{4}{*}{GPT-5.2}
  & direct        & 0.303          & 0.025          & \textbf{0.597} & 0.779 \\
  & codex         & 0.343          & 0.000          & 0.464          & \textbf{0.819} \\
  & terminus-2    & 0.514          & 0.188          & 0.531          & 0.781 \\
  & terminus-kira & \textbf{0.600} & \textbf{0.363} & 0.569          & 0.713 \\
\bottomrule
\end{tabular}
\caption{Codex CLI, Terminus-2, Terminus-Kira, and Claude Code harnesses on DeonticBench. Accuracy columns report exact-match accuracy; Macro F1 columns report macro-averaged F1. Best performance is \textbf{bolded}.}
\label{tab:codex-cc-vs-terminus}
\end{table*}

\subsection{Claude Code and Codex CLI}
\label{appendix:additional-cc-codex}
Table~\ref{tab:codex-cc-vs-terminus} extends the main comparison with Codex CLI and Claude Code. We omit a (model, harness) cell when the harness does not
natively support the model. The setup is the same as described in section \ref{setup-harness}.

\paragraph{Claude Code is a strong scaffold for open-source Qwen models.} On the Qwen models for which we have a Claude Code run (Qwen3 and Qwen 3.5 models), Claude Code delivers the highest SARA-Numeric accuracy on three of four, with Qwen3.5-397B as the exception where Terminus-KIRA performs better. Claude Code is also the only harness that recovers non-trivial Airline accuracy on open-source models: 0.050--0.113 across the
four Qwens, compared with the near-zero numbers we see under Codex and Terminus-2. The gain is concentrated on the \emph{numerical} tasks; on the classification tasks Claude Code is competitive but not consistently dominant. However, \textbf{direct prompting remains a strong baseline that Claude Code does not uniformly beat for weaker models.}

\paragraph{Codex CLI is a comparatively light-weight scaffold.} For most models in the table, Codex produces lower SARA-Numeric accuracy than the other harnesses available for the same model, and its Airline accuracy on open-source models is at or near zero. We interpret this as Codex adding relatively little structure on top of the underlying model on the numerical tasks: behavior under Codex stays close to direct prompting. On the classification tasks (SARA-Binary, USCIS-AAO) Codex is
broadly competitive with Terminus-2.

\paragraph{Terminus-KIRA is the strongest harness for frontier models and the largest open-weight models.}
Under Terminus-KIRA, GPT-5.2 reaches 0.600 SARA-Numeric and 0.363 Airline, well above its Codex and Terminus-2 numbers. The same ordering holds for Kimi-K2, where Terminus-KIRA is the best harness on all four tasks, and for Qwen3.5-397B, where Terminus-KIRA wins SARA-Numeric, SARA-Binary, and USCIS-AAO by large margins. For the smaller open-source models (Qwen3.5-35B and Qwen3-Coder-480B), the extra agentic
capacity that Terminus-KIRA provides does not translate into higher accuracy.

\begin{table*}[t]
\centering
\renewcommand{\arraystretch}{1.25}
\setlength{\tabcolsep}{4pt}
\begin{tabular}{ll cc c}
& & \multicolumn{2}{c}{\textbf{Accuracy}} & \multicolumn{1}{c}{\textbf{Macro F1}} \\
\cmidrule(lr){3-4} \cmidrule(lr){5-5}
Model & Setup & SARA-Num & Airline & SARA-Bin \\
\midrule
\multirow{3}{*}{GPT-5.1}
  & Direct Solving    & 0.543          & 0.863          & 0.700          \\
  & Terminus-Kira     & \textbf{0.692} & \textbf{0.889} & \textbf{0.748} \\
  & DSPy RLM          & 0.114          & 0.125          & 0.683          \\
\midrule
\multirow{3}{*}{Qwen3-Coder-480B}
  & Direct Solving    & \textbf{0.249} & \textbf{0.021} & 0.591          \\
  & Terminus-Kira    & 0.143          & 0.013          & \textbf{0.766} \\
  & DSPy RLM          & 0.029          & 0.000          & 0.697          \\
\midrule
\multirow{3}{*}{Qwen3.5-122B}
  & Direct Solving    & \textbf{0.370} & \textbf{0.150} & 0.753          \\
  & Terminus-Kira    & 0.200          & 0.050          & \textbf{0.823} \\
  & DSPy RLM       & 0.171          & 0.0375          & 0.661          \\
\bottomrule
\end{tabular}
\caption{Recursive Language Model variants compared against direct prompting and the Terminus-Kira agentic harness on DeonticBench. Accuracy columns report exact-match accuracy on SARA-Numeric and Airline; the Macro F1 column reports macro-averaged F1 on SARA-Binary. Best per (model, metric) row group is \textbf{bolded}.}
\label{tab:rlm-vs-direct-vs-kira}
\end{table*}

\subsection{Recursive Language Models}
\label{appendix:additional-rlm}

Table~\ref{tab:rlm-vs-direct-vs-kira} compares direct prompting, the Terminus-Kira agentic harness, and a Recursive Language Models setup \citep{zhang2025recursive} implemented with DSPy \citep{khattab2024dspy}. Specifically, we set the supervisor and worker to be the same model being evaluated, with a maximum of 10 iterations and 50 worker calls.

As shown in Table~\ref{tab:rlm-vs-direct-vs-kira}, RLMs significantly degrade performance on SARA-Numeric and Airline. The RLM variant is the weakest setting for every model, and the effect is most severe where the base model is strongest. For GPT-5.1, Airline accuracy drops from $0.863$ under direct prompting and $0.889$ under Kira to $0.125$ under DSPy RLM, while SARA-Numeric drops from $0.692$ to $0.114$. Qwen3-Coder-480B exhibits the same pattern at a lower absolute scale.

On the closed-class SARA-Binary task, RLMs hold up relatively well. For Qwen3-Coder-480B, DSPy RLM scores $0.697$, outperforming direct prompting ($0.591$) by $+0.11$. For GPT-5.1, DSPy RLM ($0.683$) is within $0.02$ of direct prompting ($0.700$). The exception is Qwen3.5-122B, for which DSPy RLM underperforms both other settings.

\section{Error Analysis}
\label{appendix:error-analysis}
Table~\ref{tab:error-analysis-by-category} reports the per-trial error rate of each (harness, model category) pair on the DeonticBench, broken down by failure mode. We group models into two categories: \emph{open-source}, comprising the Qwen and Kimi families, and \emph{closed-source}, comprising GPT-5.1, GPT-5.2, and Claude Sonnet 4.5. For every harness, we report the overall error rate (\textbf{Err\%}) alongside the per-type incidence of the three failure modes we observe in practice: agent timeouts (\textbf{Timeout}), harness runtime errors (\textbf{Runtime}), and output parsing failures (\textbf{ParseFail}). Each error-type entry is given as a raw count together with its share of trials in that row, so the three percentages sum (up to rounding) to the \textbf{Err\%} column.

 \textbf{Timeout} (\texttt{AgentTimeoutError}) occurs when the agent exceeds the 10 minutes budget allotted to a trial. Typically the model is still emitting tokens or the agent loop is still issuing tool calls when the time limit is reached and still no answer is produced. A \textbf{Runtime} error is raised by the harness itself when its internal machinery fails independently of the model's output; in Terminus-2, for instance, this surfaces when the agent cannot drive its underlying \texttt{tmux} shell session (e.g., a failed \texttt{send-keys} or a broken session invariant), indicating that the harness, rather than the model, was unable to make progress. A \textbf{ParseFail} occurs when the model returns a response in the wrong shape---missing the expected tool call, malformed JSON, or an answer that does not match the benchmark's required output format---so the harness cannot extract a usable prediction.

As shown in Table~\ref{tab:error-analysis-by-category}, closed-source models are remarkably reliable across every harness: their aggregate error rate is only $0.7\%$, with no runtime or parsing failures and a handful of timeouts confined to the Terminus-Kira harness. Second, the open-source category is roughly seventeen times more error-prone in aggregate ($12.1\%$), and its failures are overwhelmingly timeouts ($10.6\%$ of all trials), with parsing failures a distant second ($1.5\%$) and runtime errors essentially negligible. Third, the cost of running open-source models is strongly harness-dependent: Terminus-2 keeps the open-source error rate to $3.6\%$, codex roughly triples that at $11.8\%$, and Terminus-Kira pushes it to $27.8\%$, suggesting that the harness loop and its timeout budget interact with model latency far more than with model capability. Taken together, these results indicate that the bulk of observed instability in our experiments stems from open-source models exceeding harness time limits rather than from intrinsic agent or model failures.

\begin{table*}[t]
\centering
\setlength{\tabcolsep}{4pt}
\begin{tabular}{llrrrr}
\toprule
\textbf{Harness} & \textbf{Category} & \textbf{Err\%} & \textbf{Timeout} & \textbf{Runtime} & \textbf{ParseFail} \\
\midrule
\multirow{2}{*}{codex}          & open-source   & 11.8\% & 86 (9.9\%)   & 0 (0.0\%) & 16 (1.8\%) \\
                                & closed-source & 0.6\%  & 1 (0.6\%)    & 0 (0.0\%) & 0 (0.0\%)  \\
\midrule
\multirow{2}{*}{Terminus-Kira}  & open-source   & 27.8\% & 239 (24.9\%) & 1 (0.1\%) & 26 (2.7\%) \\
                                & closed-source & 1.7\%  & 7 (1.7\%)    & 0 (0.0\%) & 0 (0.0\%)  \\
\midrule
\multirow{2}{*}{Terminus-2}     & open-source   & 3.6\%  & 53 (3.0\%)   & 0 (0.0\%) & 10 (0.6\%) \\
                                & closed-source & 0.0\%  & 0 (0.0\%)    & 0 (0.0\%) & 0 (0.0\%)  \\
\midrule
\multirow{2}{*}{\textbf{Total}} & open-source   & 12.1\% & 378 (10.6\%) & 1 (0.0\%) & 52 (1.5\%) \\
                                & closed-source & 0.7\%  & 8 (0.7\%)    & 0 (0.0\%) & 0 (0.0\%)  \\
\bottomrule
\end{tabular}
\caption{Error breakdown by harness and model category. Open-source covers Qwen and Kimi models; closed-source covers GPT-5.1, GPT-5.2, and Claude Sonnet 4.5. Each error-type column shows the count and its share of trials in that row. Open-source failures are overwhelmingly timeouts, while closed-source models are essentially error-free across all three harnesses.}
\label{tab:error-analysis-by-category}
\end{table*}

\end{document}